\documentclass{article}
\usepackage{nopageno}
\usepackage{bm}
\usepackage{geometry}
\usepackage{url}
\usepackage[english]{babel}
\usepackage{setspace}
\usepackage{graphicx}
\usepackage{color}
\usepackage{rotating}
\usepackage{subcaption}
\usepackage{kbordermatrix}
\usepackage{amsmath}
\usepackage{tabularx}
\usepackage[longnamesfirst,sort]{natbib} 
\usepackage{comment}
\includecomment{comment}

\usepackage{tipa}

\usepackage{fancyhdr} 
\usepackage[unicode=true,pdfusetitle,
 bookmarks=true,bookmarksnumbered=true,bookmarksopen=true,bookmarksopenlevel=2,
 breaklinks=true,pdfborder={0 0 1},backref=false,colorlinks=true,citecolor=blue,urlcolor=blue,linkcolor=blue]
 {hyperref}
\hypersetup{pdfstartview={XYZ null null 1}}

\begin{document}

\fancypagestyle{firststyle}
{
   \fancyhf{}
  \lhead{Estonian word naming}
   \fancyhead[R]{\thepage}
   \renewcommand{\headrulewidth}{0pt}
}
\thispagestyle{firststyle}

\vspace*{2\baselineskip}

\begin{center}
  {\bf An experimental and computational study of an Estonian single-person word naming} \\
\ \\
\ \\
Kaidi L\~oo\\
University of Tartu, Estonia\\
\ \\
Arvi Tavast\\
Institute of Estonian Language, Estonia\\

\ \\
Maria Heitmeier\\
University of  T\"ubingen, Germany\\
\ \\
R. Harald Baayen\\
University of T\"ubingen, Germany \\
\ \\
\ \\

\vspace*{0.6\baselineskip}
Version: September 3, 2025\\
\vspace*{0.2\baselineskip}
\end{center}

\vspace*{0.6\baselineskip}

\begin{flushleft}

Corresponding author: \\
Kaidi L\~oo \\
Institute of Estonian and General Linguistics, University of Tartu\\
Jakobi 2-405, 50090 Tartu, Estonia\\
e-mail: kaidi.loo@ut.ee\\
\end{flushleft}

\newpage

\fancyhf{}
\lhead{Estonian word naming}
\fancyhead[R]{\thepage}
\renewcommand{\headrulewidth}{0pt}
\vspace*{2\baselineskip}

\begin{center}

  {\bf Abstract}
  \end{center}

\noindent
This study investigates lexical processing in Estonian. A large-scale single-subject experiment is reported that combines the word naming task with eye-tracking. Five response variables (first fixation duration, total fixation duration, number of fixations, word naming latency, and spoken word duration) are analyzed with the generalized additive model. Of central interest is the question of whether measures for lexical processing generated by a computational model of the mental lexicon (the Discriminative Lexicon Model, DLM) are predictive for these response variables, and how they compare to classical predictors such as word frequency, neighborhood size, and inflectional paradigm size.  Computational models were implemented both with linear and deep mappings. Central findings are, first, that DLM-based measures are powerful predictors for lexical processing, second, that DLM-measures using deep learning are not necessarily more precise predictors of lexical processing than DLM-measures using linear mappings, third, that classical predictors tend to provide somewhat more precise fits compared to DLM-based predictors (except for total fixation duration, where the two provide equivalent goodness of fit), and fourth, that in the naming task lexical variables are not predictive for first fixation duration and the total number of fixations. As the DLM works with mappings from form to meaning, the predictivity of DLM-based measures for total fixation duration, naming latencies, and spoken word duration indicates that meaning is heavily involved in the present word naming task. \\ \ \\

\newpage

\section{Introduction}

Classical printed dictionaries of languages such as English and Estonian present the vocabulary of a language by means of a list alphabetically ordered by lexeme. The user first has to look up this lexeme. Once located, the dictionary provides further information on inflectional variants, pronunciation, and a description of its meaning and senses.  The classical dictionary has a long history of inspiring theories of speakers' knowledge of the words and expressions of their language, a body of knowledge referred to as their `mental lexicon'.  In comprehension, the first stage of lexical processing is taken to be a process of `lexical access', resulting in the identification of the entry of the lexeme on the basis of its form. Once this entry in the mental lexicon has been identified, lexical access is completed, and information about pronunciation and semantics becomes available.  

The process of lexical access has been conceptualized in different ways. Moving away from an alphabetical ordering of form entries, \citet{Forster:76} proposed a mental dictionary with entries ordered by decreasing frequency, in order to capture the pervasive effect of frequency of occurrence \citep{Kliegl:2004, Brysbaert:etal:2011}. Alternatively, the interactive activation model of \citet{McClelland:Rumelhart:81} does not impose an ordering on the entries (restricted to word units), but assigns words a resting activation level proportional to their frequency of use.  In order to deal with morphologically complex words within these frameworks, decomposition of complex words into their constituents was introduced as this would reduce memory load and speed up processing.  \citet{Taft:Forster:76}'s affix stripping model was inspired by hash coding as described in \citet{Knuth:73}, although it is questionable whether a processing model exposed to actual word tokens would benefit from the stripping procedure \citep{schreuder1994prefix}.  For interactive activation models, which need exponentially more connections as the number of word units increase, decomposition of complex words into their constituents keeps the number of connections and the interactive competition process, feasible (but see  \citealt{Snell:2024} for possible shortcuts).   

A substantial body of literature has argued for obligatory morphological decomposition \citep[e.g.,][]{Rastle:Davis:New:2004,taft:2004,rastle2008morphological,smolka2014verstehen}.  However, \citet{Marelli:Amenta:Crepaldi:2014,amenta2015fruitless} have shown that many experimental effects are confounded with measures of orthography-to-semantics and phonology-to-semantics consistency.  \citet{Loo:2022} suggests that phenomena previously attributed to morphological decomposition may actually emerge already early on from an interplay between formal and semantic effects. Furthermore, the lexicon as a static repository of lexical knowledge has been criticized by \citet{Libben:2022}, who argued for a `flexicon', conceptualized as a dynamic and interconnected
system of symbolic units that is continuously updated with experience.
A more flexible but very different approach to lexical processing had already been developed 20 years earlier  \citep{seidenberg2014quasiregularity,Harm:Seidenberg:1999,Harm:Seidenberg:2004}, using deep learning networks. However, after the inconclusive end to the connectionist debate \citep[see][]{McClelland:Patterson:2002a,McClelland:Patterson:2002b,Pinker:Ullman:2002,Pinker:Ullman:2002b},  many researchers interested in the mental lexicon felt that connectionist models were just not up to the task, and that computational modeling is a fruitless enterprise. Discussions of experimental results within the conceptual framework of the interactive activation model flourished, and continue to flourish.    

However, a novel computational framework that builds on the insights of the connectionists, while aiming to keep models  simpler and better interpretable, has been developed: the Discriminative Lexicon Model \citep[DLM,][]{Baayen:Chuang:Shafei:Blevins:2019,Heitmeier:Chuang:Baayen:2025}. This modeling framework, implemented in the \textbf{JudiLing} package for julia \citep{Luo:2024}, embeds words' forms in a high-dimensional form space, and makes use of distributional semantics \citep{Harris:1954,Firth:1968,Landauer:Dumais:1997} to embed words' meanings in a second high-dimensional space. Central to this approach to the mental lexicon are the comprehension mappings from the form space to the semantic space, and the production mappings from the semantic space to the form space. Linear mappings are often of excellent quality, but deep mappings can also be set up. The advantage of a linear mapping is its interpretability.  For instance, when using letter 3-grams to embed words in form space, a linear mapping will associate each 3-gram with its own embedding.  The 3-grams that overlap with a given affix are correlated with the centroid of the embeddings of the words sharing that affix, i.e., with its prototypical meaning is semantic space \citep{Baayen:Berg:Mohamed:2025}. Deep mappings, on the other hand, may offer improved precision. The DLM has been shown to be able to `understand' and `produce' complex words across a wide range of typologically different languages, and also provides a series of measures that are predictive for reaction times in the visual lexical decision task \citep{heitmeier2023trial}, spoken word duration \citep{Gahl:Baayen:2024}, and Mandarin pitch contours \citep{Chuang:Bell:Tseng:Baayen:2025,lu2025realization}. 

The purpose of the present study is to test the predictions of the DLM against behavioral measures of lexical processing in Estonian.
We compare widely-used measures of lexical processing such as word frequency, orthographic neighborhood size, and word length, but also the more recently developed measure of inflectional paradigm size (introduced below) with measures obtained from the DLM framework (such as target correlation and contextual independence, introduced below).  The predictivity of these measures is studied for eye movements, word naming latencies, and acoustic durations collected from a large single person word naming study with eye-movement registration. 

In what follows, we first provide some background motivating the use of the word naming task and a single-subject mega-study design. We then introduce the discriminative lexicon model. This is followed by statistical analyses, where for each response variable, we compare generalized additive mixed models \citep{Wood:2017} with the classical predictors with the corresponding models in which the classical predictors are replaced by predictors motivated by the DLM. Our conclusions are presented in the general discussion section.  

\section{Task, predictors, and experimental design}

For probing lexical processing, the lexical decision task as well as eye-movement registration have been widely used \citep[see, e.g.,][]{Whaley:78,Balota:Cortese:Pilotti:1999,rayner1986lexical,Bertram:Hyona:2003}. The lexical decision task has been an excellent tool for gauging many aspects of lexical processing, but it is a task with a word-nonword discrimination component \citep{Balota:84,Grainger:Jacobs:96} that is not part of daily language use. 

In the present study, we utilize a different task, the word naming task, to assess the speed with which words presented on a screen can be read out loud. We combined this task with eye-tracking to clarify what measures are predictive for the earliest stages of information processing during reading. The word naming task, because it does not have a decision component, is arguably closer to normal language use. However, this task has not only its pros but also its cons.  The task combines comprehension (reading) and production (pronouncing the word). This makes it more difficult to assess what the task actually measures.  This is why we also tracked the eyes during reading, in the hope that this will provide information specifically about the initial comprehension processes involved in the task.  

Several studies have reported that semantic predictors are less predictive in the word naming task as compared to the lexical decision task \citep{Balota:etal:2004, Yap:etal:2012,Baayen:Feldman:Schreuder:2006,Baayen:Wurm:Aycock:2007}. It is possible that the naming task is carried out primarily by means of converting an orthographic representation into a phonological representation \citep{Coltheart:Rastle:Perry:Langdon:Ziegler:2001}, and if production then proceeds purely on the basis of this phonological representation, then little or no effects of meaning are expected.  As meaning plays a central role in the normal comprehension and production processes as formalized in the DLM, the naming task is especially suited for testing the predictions of this model against experimental data.
If meaning is indeed not accessed in the word naming task, then the meaning-based predictions of the DLM are predicted not to have any explanatory value for the response variables in this task.

Classical predictors for word naming latencies include word frequency, word length, and neighborhood density \citep{Coltheart:77,Yarkoni:Balota:Yap:2008}. Within interactive activation frameworks, word frequency determines resting activation levels, word length determines the amount of bottom-up input, and neighborhood density gauges the amount of between-word lexical competition.  Orthographic depth has also been reported to predict naming latencies \citep{Balota:etal:2006}. 
\citet{Loo:etal:2018b} conducted a multi-participant study of word naming with inflected nouns in Estonian, the language under investigation in the present study. They found that nouns with higher word frequency, larger orthographic neighborhoods, larger inflectional paradigms, and larger morphological families were processed with shorter naming latencies and acoustic durations.

More recently, it has also become clear that the phonetic detail of speech production is co-determined by semantics. \citet{plag2015homophony,Plag:Homann:Kunter:2017} observed that the duration of syllable-final [s] in English is co-determined by the semantics realized by this exponent (singular on verbs, plural on nouns, genitive singular or plural, etc (see also \citet{Loo:2023} for similar findings for syllable-final [s] in Estonian). \citet{Gahl:Baayen:2024} showed that the spoken word duration of English heterographic homophones is in part predictable from their semantics, using distributional semantics and the DLM model. \citet{Chuang:Bell:Tseng:Baayen:2025} and \citet{lu2025realization} document that the phonetic detail of how the tones of Mandarin Chinese word tokens in a corpus of spontaneous conversations are realized co-varies systematically with the corresponding contextualized embeddings. Clearly, word embeddings have opened up new ways of assessing systematicities in semantic space, and systematicities between form and meaning, that are outside of the scope of standard symbolic approaches to semantics \citep[see also][]{Mandera:etal:2017, Westbury:Wurm:2022,amenta2017sound,amenta2015fruitless}.

\begin{figure}
    \centering
    \includegraphics[width=0.5\linewidth]{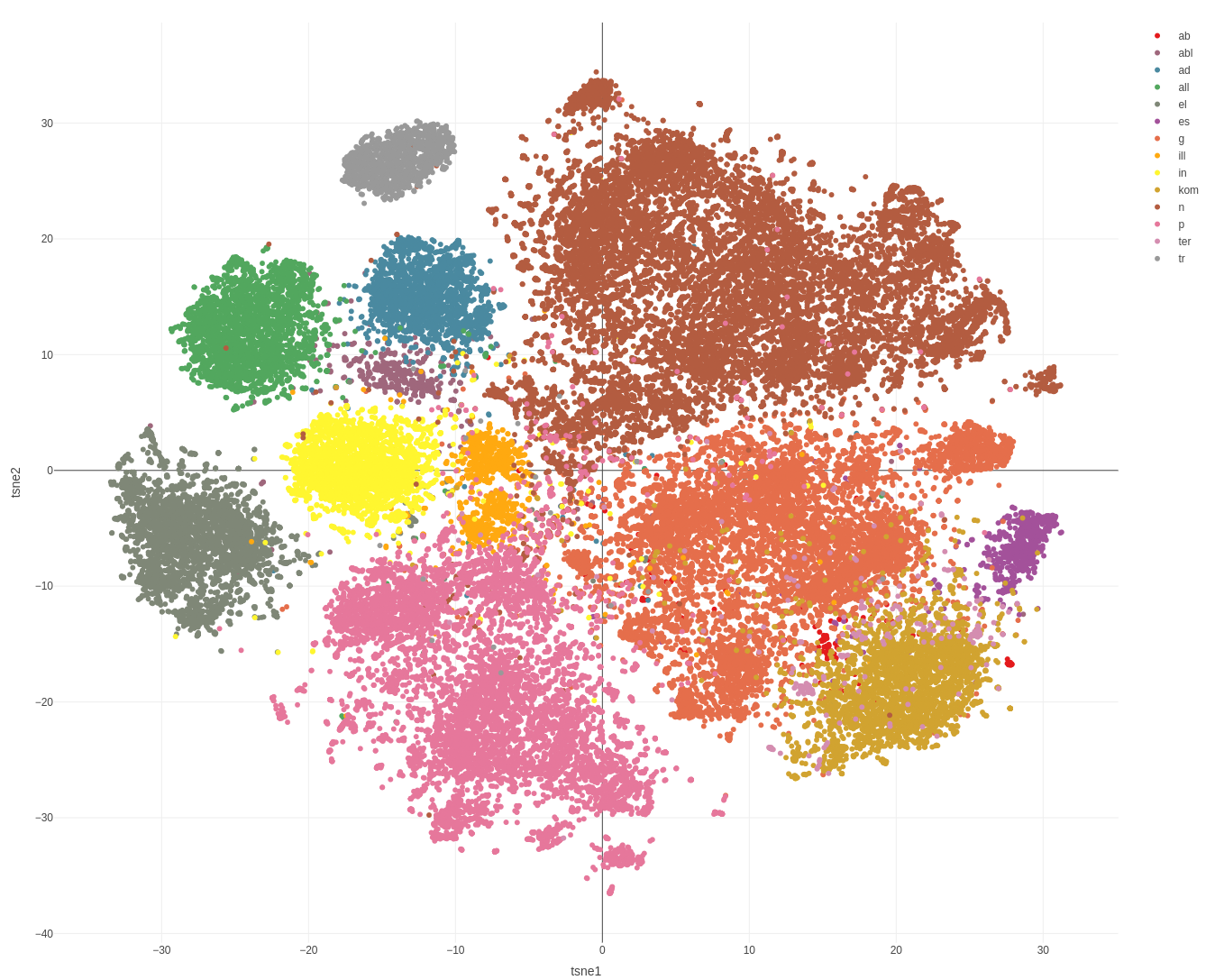}
    \caption{Two-dimensional t-SNE plot for the shift vectors of 65,404 Estonian case-inflected nouns. Shift vectors cluster by case.}
    \label{fig:Estonian_shift_vectors}
\end{figure}

For the present study, it is especially important to keep in mind that investigations of inflectional and derivational semantics using embeddings have shown that the `meanings' of classical symbolic features such as \textsc{tense}, \textsc{number} and \textsc{case} are far more intricate than was previously assumed. \citet{Baayen:Moscoso:2005} showed that in English, irregular verbs and regular verbs are positioned differently in the semantic space. \citet{shafaei2023,shafaei2024semantic} documented for English that the shift in semantic space from a singular noun to its corresponding plural noun varies systematically with semantic class (e.g., fruits, vehicles, animals).  For Russian \citep{chuang2023paradigm} and Finnish \citep{nikolaev:2022}, the shift in semantic space from a singular noun to its plural varies consistently by case. For Estonian, plural shift vectors likewise cluster by case, as shown in Figure~\ref{fig:Estonian_shift_vectors}. In order to do justice to the semantic complexities of inflected words, it is essential to work with empirical embeddings \citep[see also][]{Chuang:Bell:Tseng:Baayen:2025}.

A problem with many earlier studies of lexical processing is that they make use of underpowered experimental designs with repeated measures for subjects and items  \citep{Westfall:Kenny:Judd:2014}. Studies with a large number of words and a large number of participants overcome these problems \citep{Balota:etal:2007,Ferrand:etal:2010,Keuleers:etal:2012,Tucker:etal:2019}. Collecting data for such mega-studies is relatively expensive and time consuming, however. One solution is to run an experiment with a large number of words, but a single participant. Such a study design allows us to also use thousands of items with a relatively low acquisition cost. 

Sampling a single speaker from a language may seem nothing more than a practical solution to minimize research costs, but it is worth taking into account that it is far from clear how to assess the question of `lexical processing of speakers of a given language', in our study, Estonian. 

First, standard experiments probe the knowledge of a set of  speakers of a language, but is this knowledge in any way representative of `the language'?   Possibly, socially accepted normative Estonian (as documented in standard dictionaries and grammars) could be viewed as the emergent cumulation of the knowledge that individual speakers of Estonian have of their language. However, it is unlikely that any single speaker has full mastery of the cumulative knowledge that is represented in large corpora and dictionaries. 

A corpus can be seen as a practical approximation of the social norm, albeit limited by accessibility of texts and rendered static by its own collection process. But the knowledge captured by large corpora is far more extensive than that of any single speaker. Whereas most high-frequency words are known and used by all speakers, knowledge of low-frequency words is much more fragmentary, and depends on speaker's individual expertise, hobbies, and interests.  As a consequence, corpora overestimate both the number of words that individual speakers know, and the frequencies of the less common words in the usage of individual speakers.  This is an important caveat for the present modeling study, in which we make use of corpus-derived frequencies and type counts (when estimating orthographic neighborhood density) to probe the mental lexicon of an individual speaker of Estonian. It should be kept in mind that the size of inflectional paradigms was also calculated from a corpus, and is an estimate of community usage, and as a consequence unavoidably overestimating the paradigm sizes that characterize the active use of paradigms for our participant.

Complementing corpora, dictionaries constitute yet another window on the social norm. Modern dictionaries are corpus-based, but still mostly human-curated, discrete, smaller in volume than corpora, and even more static. Instead of representing an omniscient external observer, dictionaries are compiled by humans with their inherently limited knowledge, albeit equipped with tools and access to information well above the average speaker. For this experiment, dictionaries were used to derive  a measure of entropy, which we quantified as the uncertainty associated with the possible pronunciations of a given written word form.  The resulting measure reflects a social norm that may diverge from the knowledge of the speaker that participated in our experiment. 

Second, lexical knowledge is not static, but is updated continuously at both small and large time-scales.  For the continuous updating of lexical knowledge within large lexical decision experiments, see \citet{heitmeier2023trial,Heitmeier:Chuang:Baayen:2025}, and for continuing vocabulary development over the lifetime, see \citet{Keuleers:2015}; for the mental lexicon as a mental `flexicon', see \citet{Libben:2022}.  In this study, we work with predictors that are static in the sense that they do not take into account that the participant's mental lexicon was likely changing over time as the experiment proceeded in time.  However, we will address the question of whether in the course of the experiment, the effect of predictors change. Such changes would be indicative of changing task strategies, a changing mental lexicon, or both. 

Currently, single-subject study designs are still quite rare in psycholinguistic research \citep{Pham:2014, Sun:2018}, but have a long history in applied sciences, such as applied education, psychology, and linguistics \citep[see, e.g.,][]{Mcreynolds:etal:1986, Beeson:etal:2006, Horner:etal:2005}. Single-subject mega-studies offer the advantage of eliminating participant as a source of noise.  This simplifies statistical evaluation and makes it possible to investigate substantial numbers of different words. Compared to experiments conducted with smaller numbers of words, but larger numbers of subjects sampled from the same population, the power to detect effects of word properties increases \citep{Pham:2014}. Of course, if the interest is in the variation between speakers, designs with multiple speakers are preferable. As shown by \citet{heitmeier2023trial}, in the British Lexicon project \citep{Keuleers:etal:2012}, the effect of neighborhood density, for instance, shows enormous between-subject variability, with facilitation for some, inhibition for others, and for yet others, no effect at all. This variability exists for participants sampled from university students and staff, and thus likely underestimates substantially the variability in the lexical processing of in the highly variegated population of speakers of English. Obviously, between-subject differences in lexical processing cannot be traced with a single-subject design. For the present study, which addresses the computational modeling of lexical processing in Estonian, a single-subject design with a large number of words is preferable, not in the least as lexical resources representing the prior lexical knowledge of individual participants are not available.

\section{Modeling with the DLM}

We use the DLM to gauge one aspect of the lexical processes involved in word naming, namely, how precisely a given word's meaning can be predicted from its form. Here, we make the assumption that in the naming task, the meanings of words are engaged and are actually the starting point for the production process. As mentioned previously, the DLM conceptualises lexical comprehension as a mapping from a word's form to its meaning, and production as a mapping from meaning to form. Forms and meanings are represented as high-dimensional vectors. Form vectors are usually binary vectors indicating the presence and absence of overlapping letter n-grams in the word form. Semantic vectors are generally taken from word embedding spaces such as word2vec \citep{mikolov2013efficient}, fastText \citep{grave2018learning} or GloVe \citep{pennington2014glove}. To model a lexicon, the form and semantic vectors are collected into two matrices: the $\mathbf{C}$ matrix containing all form vectors and the $\mathbf{S}$ matrix containing all semantic vectors (example matrices can be seen in Figure~\ref{fig:C_and_S}). For modeling the transformation between meaning and form, various methods have been explored, and are implemented in the \textbf{JudiLing} package  \citep{Luo:2024}. First, in Linear Discriminative Learning (LDL), two linear transformations are computed: $\mathbf{F}$, a transformation from the form to the meaning matrix modeling comprehension, and $\mathbf{G}$, a transformation from the meaning to the form matrix modeling production:

$$\mathbf{S} = \mathbf{CF}$$
$$\mathbf{C} = \mathbf{SG}$$

\noindent
Since for larger matrices $\mathbf{C}$ and $\mathbf{S}$ the transformations are only approximate, we write:

$$\hat{\mathbf{S}} = \mathbf{CF}$$
$$\hat{\mathbf{C}} = \mathbf{SG}$$

\begin{figure}
    \centering
    \begin{subfigure}{0.5\textwidth}
        $$
\renewcommand{\kbldelim}{(}
\renewcommand{\kbrdelim}{)}
\renewcommand{\kbrowstyle}{\displaystyle}
\renewcommand{\kbcolstyle}{\displaystyle}
    \mathbf{C} = \kbordermatrix{ & \mathtt{\#cu} & \mathtt{cup} & \mathtt{up\#} & \mathtt{cut} & \mathtt{ut\#} & \mathtt{cub} & \mathtt{ub\#}\\ 
    \text{cup} & 1 & 1 & 1 & 0 & 0 & 0 & 0 \\
    \text{cut} & 1 & 0 & 0 & 1 & 1 & 0 & 0 \\
    \text{cub} & 1 & 0 & 0 & 0 & 0 & 1 & 1 \\} 
$$
\caption{Form matrix $\mathbf{C}$}
    \end{subfigure}
    
    \begin{subfigure}{\textwidth}
        $$\kbalignrighttrue
\renewcommand{\kbldelim}{(}
\renewcommand{\kbrdelim}{)}
\renewcommand{\kbrowstyle}{\displaystyle}
\renewcommand{\kbcolstyle}{\displaystyle}
    \mathbf{S} = \kbordermatrix{ ~ & \texttt{S1} & \texttt{S2} & \texttt{S3} & \texttt{S4} \cr \text{cup} & 0.37 & 0.66 & 0.61 & 0.16 \cr \text{cut} & 0.38 & 0.93 & 0.95 & 0.86 \cr \text{cub} & 0.52 & 0.53 & 0.08 & 0.9 \cr}$$
    \caption{Semantic matrix $\mathbf{S}$}
    \end{subfigure}

    \caption{Example of $\mathbf{C}$ and $\mathbf{S}$ matrices for a toy lexicon containing the words \textit{cup}, \textit{cut} and \textit{cub}.}
    \label{fig:C_and_S}
\end{figure}

\noindent
Instead of applying a linear transformation, the mapping between $\mathbf{S}$ and $\mathbf{C}$ can also be modeled with non-linear, deep learning networks \citep[Deep Discriminative Learning, DDL,][]{heitmeierEtAl2024}, in which case we replace the linear $\mathbf{F}$ and $\mathbf{G}$ matrices with two non-linear functions $f$ and $g$ representing the deep-learning networks together with two sets of trainable parameters $\mathbf{\Theta}_f$ and $\mathbf{\Theta}_g$:

$$\hat{\mathbf{S}} = f(\mathbf{C}, \mathbf{\Theta}_f)$$
$$\hat{\mathbf{C}} = g(\mathbf{S}, \mathbf{\Theta}_g)$$

\noindent
As the focus of this study is on comprehension, in what follows, we do not discuss production mappings. 

\subsection{Model implementation}

For computing $\mathbf{F}$  as well as for obtaining $\mathbf{\Theta}_f$, methods are available which generally follow  one of three scenarios for lexical learning:

\begin{enumerate}
    \item Learning of a mapping independent of frequency of use and order of acquisition. 
    \item Learning of a mapping, taking frequency into account, but disregarding order of acquisition.  
    \item Learning of a mapping, taking both frequency of occurrence and order of learning into account. 
\end{enumerate}

For the present study, we obtained the best results with two frequency-informed models, following  scenario 2, one using linear mappings (FIL), and one using incremental deep learning (FIDDL). These models do not take into account the order in which words were learned. \footnote{
Order-sensitive incremental learning with linear mappings implementing the Widrow-Hoff learning rule has been used to model second and third language acquisition \citep{chuang2021bilingual} as well as trial-to-trial learning in the British Lexicon Project \citep{heitmeier2023trial, keuleers2012british}.} 

Frequency-informed learning (FIL) implements frequency-weighted multivariate multiple regression. For technical details, the reader is referred to \citet{Heitmeier:Chuang:Axen:Baayen:2023} and \citet{Heitmeier:Chuang:Baayen:2025}. Multivariate multiple regression mappings (henceforth standard linear mappings) without frequency weights have been used in a wide range of modeling studies addressing lexical processing in, for example, Estonian, Korean, Mandarin, Kinyarwanda, Latin and English \citep{chuang2020estonian, chuang2021vector, vijver2022word, Baayen2018, baayen2019discriminative}. Whereas standard linear mappings provide type-based learning, FIL implements usage-based learning. The frequencies that FIL used for weighting were the frequencies provided by the Estonian National Corpus. \citep{Koppel:2023}.\footnote{https://doi.org/10.15155/3-00-0000-0000-0000-08C04M}
The number of parameters of the FIL model (the weights on the connections in its mapping from form to meaning) is 4,981 distinct letter trigrams $\times$ 300 fastText dimensions = 1,494,300. 

We also implemented a model with deep discriminative learning, using a training regime referred to with the acronym FIDDL (frequency-informed deep discriminative learning).  As a first step, a list of all tokens was obtained by expanding the type frequency list of the words in the experiment reported below. These frequencies were obtained from the Estonian National Corpus 2023 (ENC23) \citep{Koppel:2023}. The total number of tokens of the (19,314) word types in our dataset is too large for modeling (79,792,744). We therefore scaled down the original frequencies by a factor 1000, and rounded upwards, resulting in a type list with 19,314 words and a total of 289,580 tokens. From this list, we created a random sequence of the 289,580 tokens, with each type represented by its corresponding amount of tokens.

The FIDDL model was trained on this list of tokens with the mean squared error (MSE) loss, a single hidden layer with 1000 units, the Adam optimizer with learning rate 0.001, and batchsize 512. The number of parameters for the FIDDL model was 4981 $\times$ 1000 $\times$ 300 = 1,498 billion.

Both the FIL and FIDDL models were trained with 300-dimensional fastText vectors \citep{grave2018learning} to represent words' meanings, and with vectors specifying words' letter trigrams to represent their forms. 

\subsection{Comprehension accuracy}

Type-based accuracies were 17.4\% for FIL and 78.4\% for FIDDL.  In a type-based evaluation,  FIL struggles with low-frequency words. However, its accuracy evaluated in terms of percentage of tokens recognized correctly is substantially higher, at 74.4\%.  Token-wise evaluation of FIDDL is at 97.0\%.  

For each of these two models, we calculated a measure, \textit{TargetCorrelation}, that specifies for a given word what the Pearson correlation is of that word's predicted embedding and its gold-standard fastText embedding. The better and more precise a word's meaning has been learned, the shorter naming latencies are expected to be. Given the very different accuracies of the two models, a question of particular interest is to see which model provides the \textit{TargetCorrelation} measure that best predicts human performance. 

\section{Word Naming Experiment}

\subsection{Participant and materials}

There was a single participant for the experiment, a 45-year-old male Estonian native speaker (the second author of the paper). The materials were 30,000 isolated single Estonian words that varied in their surface frequency.

Materials were taken from the Estonian National Corpus 2013\footnote{
https:doi.org/10.15155/1-00-0000-0000-0000-00158L
} corpus. Based on the corpus, a word frequency list was created. This frequency list was divided into 30 frequency bands, and from each frequency band 1000 words were randomly selected. This list was randomized and divided into 30 experimental lists. Each experimental list contained 1000 words. 

\subsection{Experimental procedure}

The participant was instructed to read aloud the words on the computer screen as naturally as possible. He was seated in front of a computer screen with a distance of 60 cm. The participant rested his forehead on the Eyelink 1000 head rest. The chin rest was removed to allow speaking.

The experiment was programmed and presented using the ExperimentBuilder software by SR Researach. The screen size was 27 inches and the screen resolution was 1024 x 768 pixels. 

Before the experimental task, the eye-tracking system was calibrated for spatial accuracy below 0.5. Each trial started with a drift correction on the left, followed by a blank for 500 ms after which the target appeared on the screen. The stimuli were presented in the black Courier New Bold font (font size 48) on a white background for 2000 ms, followed by a blank screen for 750 ms. One session was divided into 20 blocks. Each block contained 50 items, and there was the possibility of taking a break after each block. A single session lasted approximately 90 minutes. The 30 experimental lists were run on different days, but approximately at the same time of day (starting around 10 am). Data collection was carried out over a period of half a year.

During debriefing after the experiment, the participant reported enjoying the experiment itself, not just the associated feeling of accomplishment, contrary to what might be expected from the relatively uncomfortable experiment setting (2-hour sessions of a tedious activity, with the head in a fixed position for eye-tracking). Monotonous timing of the stimuli, combined with total absence of other auditory or visual input in the dimly lit and soundproofed recording studio created a feeling akin to meditation, which the participant had some experience with and enjoyed in its own right. A less expected and more specific side effect was caused by the stimuli themselves. Selected randomly from corpus frequency bands without any regard to semantics, the single-word stimuli induced a constant stream of transitions between unrelated meanings and domains. The formation of unexpected semantic connections was also facilitated by the lack of context around the words, leaving room for creative interpretation and guided meditation. The participant reported a noticeable increase in creative thinking, including novel ideas on topics he was working on at the time. After a few initial sessions, he started bringing a notebook to the lab to write down his thoughts during breaks. At the end of the experiment, the participant not only expressed willingness to repeat the task at any time, but also recommended it as a creative thinking tool. These reflections strongly suggest that semantics does play a role in this word naming experiment. 

\subsection{Analysis}

For statistical evaluation, we used generalized additive mixed effects models (GAMMs, \citealt{hastie_generalized_1990,Wood:2017};  R-package \textit{mgcv}, version 1.8). GAMMs allow the testing of nonlinear relationships between the dependent and the predictor variables.  The analyses we present are the result of exploratory data analyses.  We set $\alpha = 0.0001$, and only report predictors with $p < \alpha$.  Only the simplest, well-supported models are reported.  

\subsection{Preprocessing}

Before the analysis, the audio and eye-tracking data was checked and pre-processed. The audio data for the first three lists were removed due to technical issues with the recording setup. One list was accidentally recorded twice, so we only included the first recording.

Word durations and production latencies were automatically calculated from the recording using a Matlab script. 478 words were identified as misspoken in this process and removed. This left us with audio information for 25523 words. 

Eye gaze data information was gained using fixation reports from Eyelink Dataviewer. Eye-tracking data was not recorded for two full lists and additional 1049 trials across the remaining lists were removed due to audio recording errors. This left us with 23,473 word responses that had both eye-tracking and word production information.

Finally, all responses with acoustic durations greater than 1.5 seconds were considered outliers (2.5 standard deviation of the mean). This left us with the final data set that contains word duration, production latency, and eye fixation information for 19,314 words.

\subsection{Response variables}

We investigated three eye-tracking measures as response variables to capture early and late reading: \textbf{First fixation duration} (the duration of the initial fixation in milliseconds), \textbf{Total fixation duration} (the total reading time on the word), and \textbf{Number of fixations} (how many fixations a word received).

Furthermore, we looked at two production measures: \textbf{Naming latency} (the time taken to start articulating the word from the time it appeared on the screen), and \textbf{Acoustic duration} (the time of the articulation). Production latencies were logarithmically transformed. Without this transformation, the model residuals depart substantially from normality.

\subsection{Predictors}

For each of the words in our dataset, we considered the following predictors.

\texttt{Word frequency} is the token frequency of the stimulus word. Previous research \citep{Loo:etal:2018a,Loo:etal:2018b} has shown that in Estonian, more frequent nouns are read faster and named faster than less frequent words, as has been observed for many other languages \citep{Whaley:78,Balota:Cortese:Pilotti:1999}. Since stimuli in word naming were presented in isolation, without any contextual information to aid in disambiguation of possible homonyms, word frequencies are cumulated for homographs. Word frequency counts are based on the ENC23, which comprises 3.8 billion tokens. Word frequency counts were log-transformed in order to avoid adverse effects of higher-frequency outliers.

\texttt{Inflectional paradigm size} is the number of forms observed for a certain lemma from the Balanced Corpus of Estonian, which comprises of 15 million tokens \footnote{https://www.cl.ut.ee/korpused/grammatikakorpus/}. For example, for the lemma \textit{senat} `senate', the paradigm size was four: \textit{senat} `nominative singular, \textit{senati} `genitive singular', \textit{senatile} `allative singular, \textit{senatis} `inessive singular'. The distribution of paradigm sizes was highly skewed. This skew was mostly eliminated by a square root transformation. Previously, inflectional paradigm size effects in Estonian have been reported in visual lexical decision task \citep{Loo:etal:2018a} and text reading \citep{Loo:2024}, as well as in word production task \citep{Loo:etal:2018b} and in spontaneous speech \citep{Loo:2023}. 

Orthographic neighborhood density \texttt{NCount} \citep{Coltheart:78} is the number of orthographic neighbors, calculated with the R-package \textbf{vwr} from the Balanced Corpus of Estonian using the dictionary form of the word. The orthographic neighborhood counts were logarithmically transformed in order to reduce the strong rightward skew in their distribution. 

Although neighborhood density has been widely studied \citep{Landauer:Streeter:73,Balota:etal:2007}, there is no general consensus on its effects on lexical processing. For English lexical decision, \citet{heitmeier2023trial} observed, for the British lexicon project \citep{keuleers2012british}, that the effect of neighborhood density varied substantially between individual participants, with no clear main effect. For Estonian, no neighborhood effect was present for visual lexical decision \citep{Loo:etal:2018a}, while for word naming, a small inhibitory effect has been reported \citep{Loo:etal:2018b}.

We also considered an \texttt{Entropy} measure for the possible pronunciations of a written word form. For example, in the word form \textit{kassi} 'cat, gen/part', the [s] is pronounced long for the genitive, and overlong for the partitive. As most of the words had unambiguous pronunciations,  we decided to make use of a simpler variable set to 0 when a single pronunciation of the form was possible, and 1 when multiple pronunciations were possible. The count of pronunciations was based on the Combined Dictionary of Estonian \citep{langemets:2024, Tavast:2021}. 

\texttt{Word Length} is the number of characters of the stimulus word. Unsurprisingly, longer words take longer to read and produce.

\texttt{POS} is the part of speech of the stimulus word. Function words tend to be pronounced with shorter acoustic duration and read faster than content words, and verbs are pronounced with shorter duration and read faster than nouns \cite[e.g.,][]{Gahl:etal:2012,Dilts:2013, Loo:etal:2024, Rayner:etal:1998,Furtner:etal:2009}. 

\texttt{Manner} of articulation of the first segment in the word, with levels approximant, fricative, nasal, plosive, trill, vowel. This factorial predictor was included to control for artifacts of the voice key, the sensitivity of which varies by manner of articulation.

\texttt{Target Correlation:} This measure provides the Pearson correlation between the predicted semantic vector and its corresponding gold standard target vector. If the target correlation for a word is higher, this implies that the DLM model has been more successful in learning the association between that word's form and its meaning. In other words, the higher the target correlation, the more accurate the mapping from form to meaning is. Previous work found that Target Correlation is predictive for reaction times in lexical decision experiments, particularly when calculated from FIL or FIDDL models \citep[e.g.][]{Heitmeier:Chuang:Axen:Baayen:2022, Heitmeier:2024}. 

\texttt{Contextual Independence:} This measure was first proposed by \citet{Gahl:Baayen:2024} as a tool for assessing the extent to which a word can be understood independently of the context in which it occurs. Whereas in normal language use, words appear in the context of other words, with meanings that are co-determined by these contexts, in the lexical decision and word naming tasks, words are presented in isolation. Participants must make decisions about these words without the help of words in the context. From a learning perspective, we can consider the target word itself and the other words with which it co-occurs as cues that predict each other.  \citet{Gahl:Baayen:2024} therefore used the Rescorla-Wagner learning rule \citep{Rescorla:Wagner:1972} to predict the words in an utterance from the words in that utterance. Denoting the number of different words by $n$, with infinite learning, the $n \times n$ matrix $\bm{W}$ predicting words from each other will converge to the identity matrix, i.e., a mapping from words to words in which any word $w$ will predict itself with weight 1, and any other word with weight 0. However, with much more restricted learning experience, words will predict themselves less well, and will depend on the words they co-occur with to be fully supported. The \texttt{Contextual Independence} measure for word $w$ is its weight (equivalently its connection strength) to itself, i.e., its diagonal value in $\bm{W}$.  \citet{Gahl:Baayen:2024} calculated the \texttt{Contextual Independence} on the basis of the British National Corpus \citep{Burnard:1995}, and found it to predict visual lexical decision latencies as well, and perhaps better, than the frequencies in this corpus. This led us to also calculate \texttt{Contextual Independence} for Estonian.  The algorithm was trained on the lemmatized ENC23 corpus\footnote{https://doi.org/10.15155/3-00-0000-0000-0000-08565L} with a learning rate $\lambda = 0.001$. Just like word frequency distributions, the distribution of contextual independence has a long right tail. The rightward skew was substantially reduced by means of a logarithmic transformation.

Conceptually, the two learning based predictors decompose the standard frequency effect into two components.  The first component, represented by Target Correlation, approximates the effect of usage on the learning of what words mean given their forms. The learning process itself is co-determined by word similarity (thus implicitly capturing neighborhood effects) and the systematic correspondences between form and meaning due to morphology (thus implicitly capturing inflectional paradigm size effects).  The second component, represented by Contextual Independence, approximates (more crudely than we would like) the consequences of the fact that words are not islands in a frequency list, but are used in utterances together with other words. In the naming task, words occur in isolation, but this does not imply that the experience gained with words in context is inert.  The contextual independence measure seeks to assess the extent to which words depend on the words in their context.  The more collocationally restricted a word is, the more it will depend on its collocates, and the lower its contextual independence will be.

Finally, we had two predictors that capture how the participant proceeded throughout the experiment.
\texttt{List nr} marks the order of the stimulus lists, ranging from 1--30. \texttt{Trial} marks the order of the stimulus within the list, ranging from 1--1000.


\subsection{Results}

\subsubsection{First fixation duration}

\begin{figure}[htbp]
    \centering
    \includegraphics[width=0.8\linewidth]{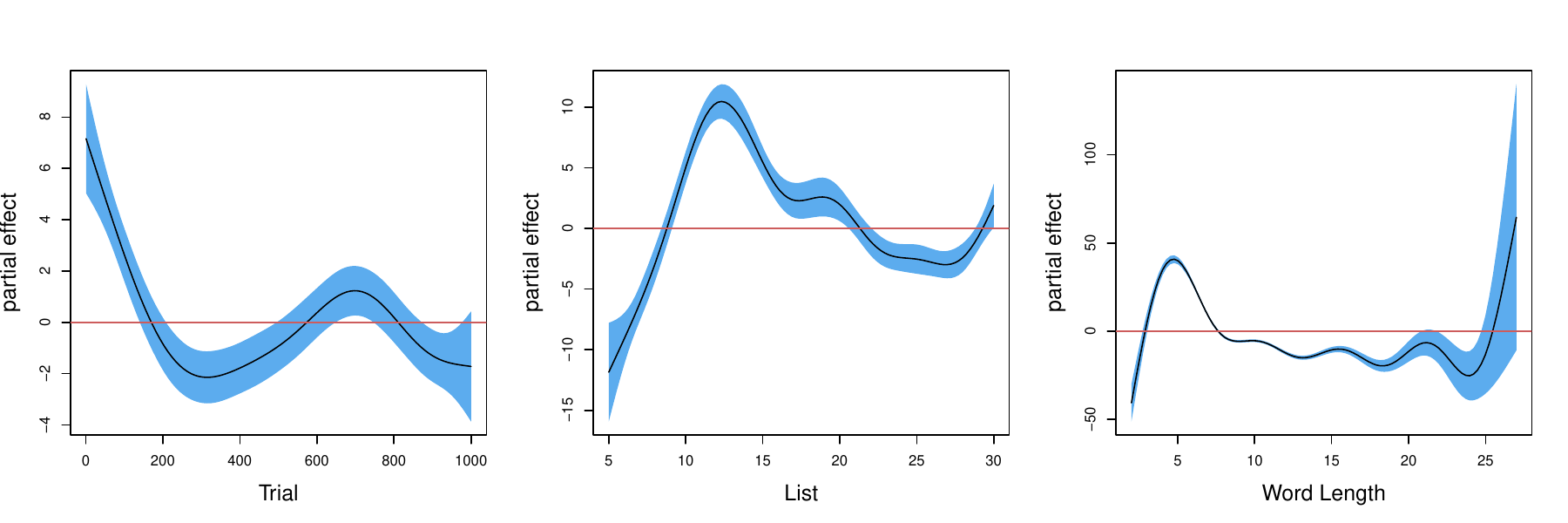}
    \caption{Partial effects in the quantile GAM fitted to the first fixation durations.}
    \label{fig:f1dur}
\end{figure}

The first fixation duration was modeled with a quantile gam \citep{fasiolo2020scalable}, as the distribution of the first fixation durations was highly non-normal and could not be corrected to normality.  Non-linear effects for median first fixation duration were present for the two time scales (\textit{Trial} and \textit{List}) and for word length. The effect of these  predictors is presented in Figure~\ref{fig:f1dur}. First fixations were the longest for words with approximately five letters. For longer words, first fixation durations may be shorter, as a subsequent fixation to a later point in the word is more likely to occur. For very long words, the median was also high, but confidence intervals are wide.  Frequency and target correlation were not predictive for first fixation durations, suggesting that the initial fixation duration gauges visual scanning.

\begin{table}[ht]
\centering
\begin{tabular}{lrrrr}
   \hline
A. parametric coefficients & Estimate & Std. Error & t-value & p-value \\ 
  Intercept & 737.6596 & 0.2739 & 2692.8836 & $<$ 0.0001 \\  \hline
B. smooth terms & edf & Ref.df & F-value & p-value \\ 
  s(Word Length) & 8.6993 & 8.9314 & 2388.7411 & $<$ 0.0001 \\ 
  s(Trial) & 5.4706 & 6.6112 & 79.2990 & $<$ 0.0001 \\ 
  s(List) & 7.9713 & 8.7150 & 342.3884 & $<$ 0.0001 \\ \hline
\end{tabular}
\caption{Summary of the quantile GAM fitted to the first fixation durations.} 
\label{tab.qgam}
\end{table}

\newpage

\subsubsection{Total fixation duration}

\begin{figure}[htbp]
    \centering
    \includegraphics[width=0.8\linewidth]{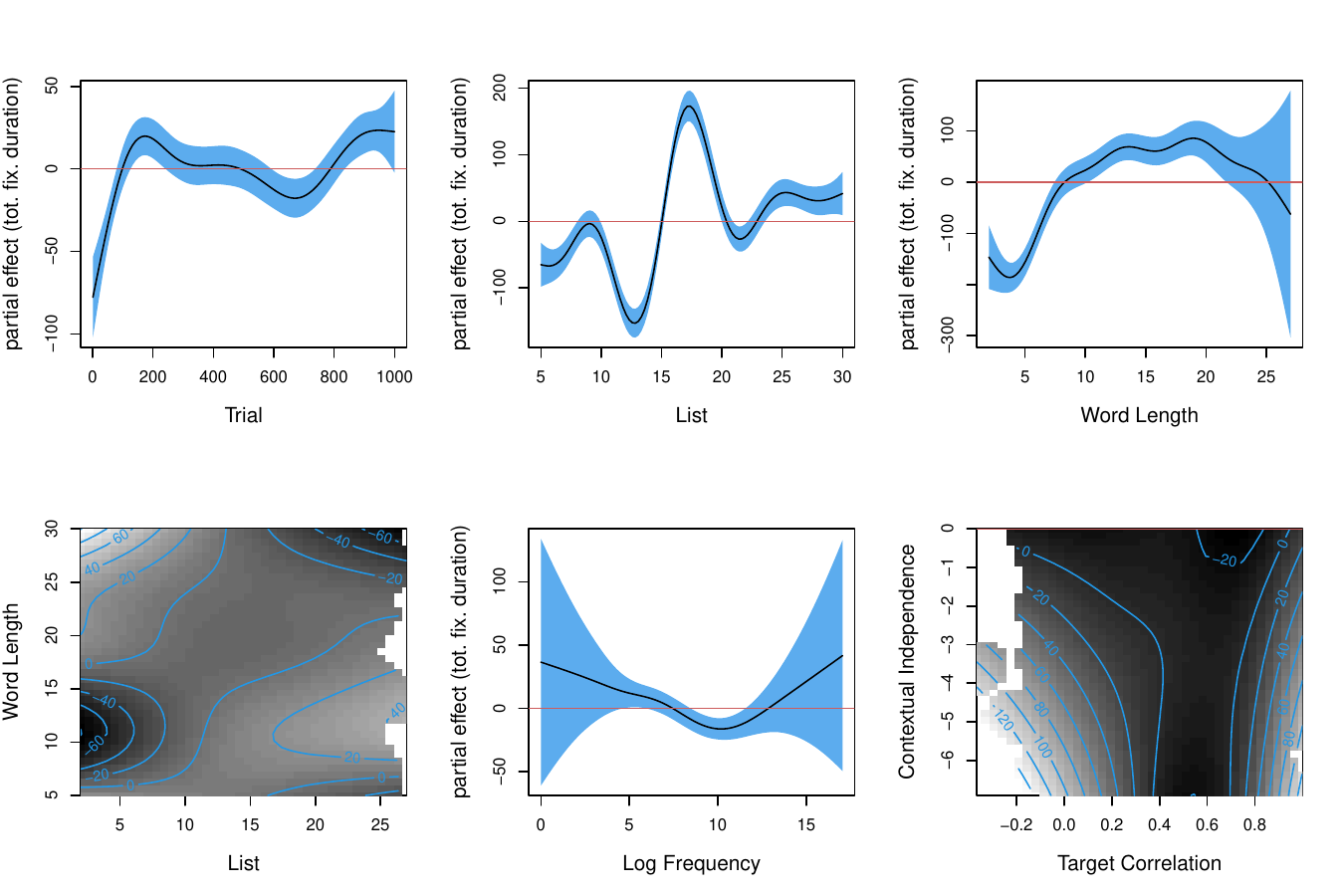}
    \caption{Partial effects in a Scaled t(3,199.917) GAM fitted to the total fixation durations with classical predictors (panels 1--5). The bottom right panel shows the partial effect of the interaction of FIL-derived  Target Correlation by Contextual Independence in the corresponding Gaussian GAM replacing Frequency.}
    \label{fig:totfixdur}
\end{figure}

Total fixation duration was modeled with a scaled-t GAM, using the \texttt{scat} directive of the \texttt{bam} function of the \textbf{mgcv} package \citep{Wood:2017}. In addition to non-linear effects of \textit{Trial} and \textit{List}, an effect of \textit{Word Length} was present, such that longer words elicited longer total fixation durations (all $p \ll 0.0001$), as expected. This effect, however, leveled off for the longer words, as shown in the upper right panel of Figure~\ref{fig:totfixdur}
The contour plot for the interaction of word length by list visualized in the lower left panel of Figure~\ref{fig:totfixdur}.
The effect of length changed as the experiment proceeded. Later in the experiment, the participant tended to fixate less long on long words, whereas early in the experiment, the long words elicited somewhat longer total fixation durations. 

Furthermore, for \textit{Frequency} a modest nonlinear effect was observed, with the expected decrease in total fixation duration for more frequent words. For the most frequent words, this trend leveled off. The AIC of this model was 271583.5.  \textit{Frequency} can be replaced by FIL-based \textit{Target Correlation} with an equivalent AIC (271584.7).\footnote{
The corresponding AICs for FIDDL was 271589.4.
}
The concurvity for the interaction of Target Correlation and Contextual Independence was 0.21, whereas the concurvity of Frequency was 0.45. The model with learning-based measures therefore provides the same goodness of fit with less confounded predictors.  

\begin{table}[ht]
\centering
{\footnotesize
\begin{tabular}{lrrrrrr} \hline
A. parametric coefficients & Estimate & Std. Error & t-value & p-value &  &  \\ 
  Intercept & 1310.3009 & 12.7881 & 102.4625 & $<$ 0.0001 \\ \hline
B. smooth terms & edf & Ref.df & F-value & p-value & $\Delta$AIC & concurvity \\ 
  s(Trial) & 6.8837 & 7.9764 & 8.1604 & $<$ 0.0001 & 111      &  0.0030      \\ 
  s(List) & 8.8918 & 8.9945 & 58.5877 & $<$ 0.0001 & 1116      & 0.9639       \\ 
  s(Log Frequency) & 3.2988 & 4.1496 & 5.6339 & 0.0001 &  15     & 0.4491       \\ 
  s(Word Length) & 7.5014 & 8.1563 & 179.1797 & $<$ 0.0001 &  882     & 0.6388       \\ 
  ti(Word Length, List) & 5.9845 & 7.6903 & 7.8129 & $<$ 0.0001 &       & 0.4772     \\ \hline
\end{tabular}
}
\caption{Summary of a Scaled t(3,199.917) GAM fitted to the total fixation durations, with classical predictors (AIC: 271,583.5).} 
\label{tab:totfixdur}
\end{table}

\subsubsection{Number of fixations}

\begin{figure}[htbp]
    \centering
    \includegraphics[width=0.6\linewidth]{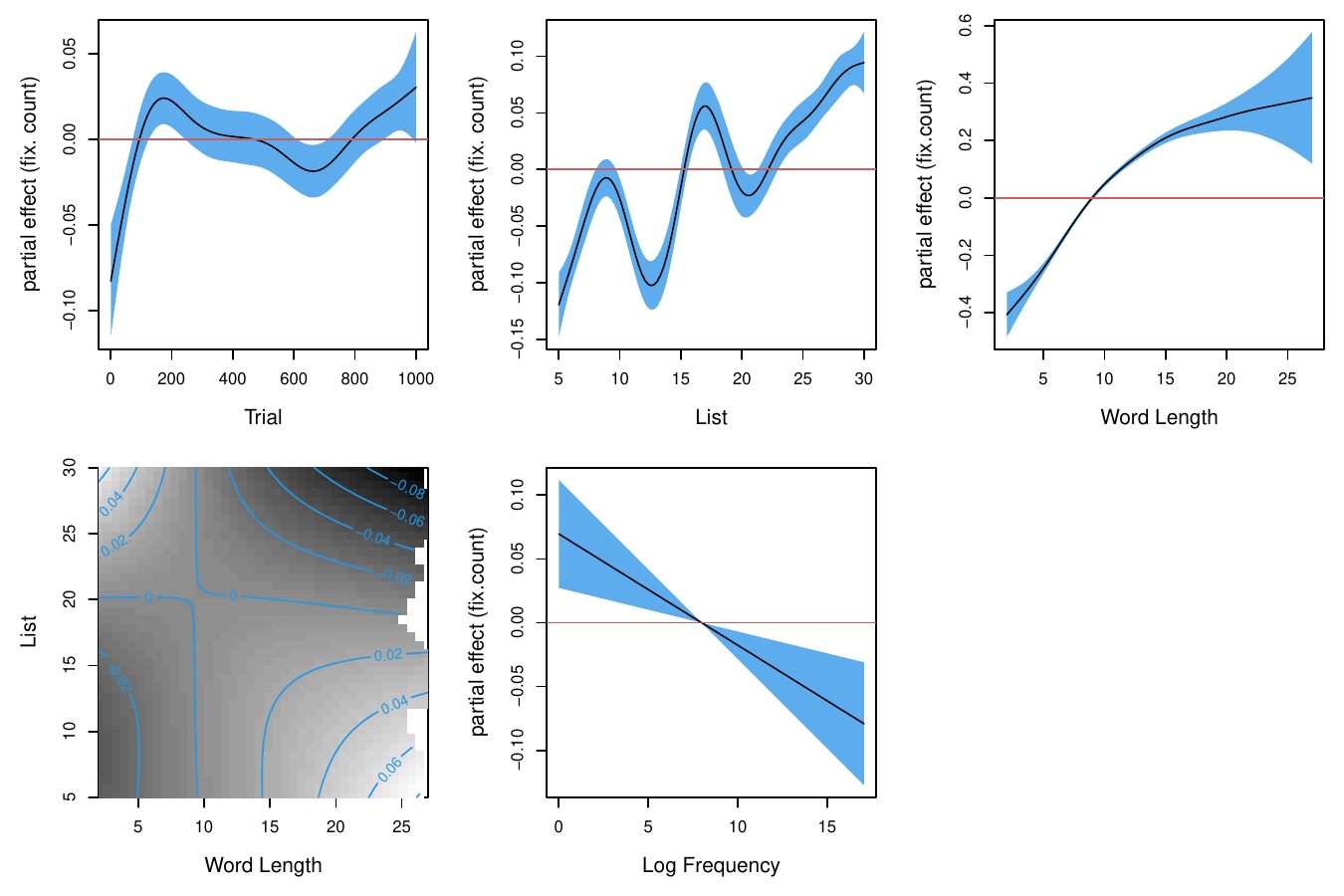}
    \caption{Partial effects of \textit{Trial}, \textit{List}, \textit{Word Length}, a interaction of \textit{List} by \textit{Word Length}, and \textit{Word Frequency} on the total number of fixations, according to a GAM with a Poisson link.}
    \label{fig:fixcount}
\end{figure}

\noindent
The number of fixations, analyzed with a Poisson GAM, varied non-linearly with \textit{Trial}, \textit{List}, and \textit{Word Length}, as shown in Figure~\ref{fig:fixcount} (all $p \ll 0.0001$).  A summary of the model is provided in Table~\ref{tab:fixcount}.

The effects of list and of trial within list show how attention changed at shorter and longer time scales. As expected, the number of fixations increased with word length. The effect of word length and list interacted. In the course of the experiment, the overall trend of list indicates that as the experiment proceeded, more fixations were executed.  The interaction of word length and list indicates that early in the experiment (low values of list), the effect of word length was stronger, and that later in the experiment (high values of list), its effect was weaker.  Apparently, in the course of the experiment, short words became progressively more difficult.  However, according to our preset $\alpha=0.0001$, it is doubtful that this  interaction would replicate.

In addition to these predictors, a linear effect was observed for \textit{Word Frequency}, such that a greater frequency of occurrence afforded fewer fixations. However, as for the interaction of word length and list, this effect is unlikely to be robust, given that we set $\alpha=0.0001$.  We also note here that the variable importance of frequency is the smallest of all predictors, and is even less than the variable importance of trial.  As frequency shows mild concurvity with word length, it is well possible that the effect of frequency (if it is really there) is at least in part an effect of length. Finally, we observed that DLM measures were not supported for the total number of fixations.


\begin{table}[ht]
\centering
\begin{tabular}{lrrrrrr}
   \hline
A. parametric coefficients & Estimate & Std. Error & t-value & p-value &  & \\ \hline
  Intercept & 1.5641 & 0.0033 & 472.0718 & $<$ 0.0001 &  \\  \hline
B. smooth terms & edf & Ref.df & F-value & p-value  & $\Delta$ AIC & Concurvity \\ \hline
  s(Trial) & 6.3709 & 7.5199 & 38.6342 & $<$ 0.0001 & 30.6 & 0.003 \\ 
  s(List) & 8.4709 & 8.9192 & 326.4415 & $<$ 0.0001 & 321.2 & 0.002 \\ 
  s(Word Length) & 4.6411 & 5.6168 & 1475.2666 & $<$ 0.0001 & 1489.6 & 0.115 \\ 
  s(Word Frequency)     & 1.0002 & 1.0004 & 10.8264 & 0.0010 & 9.86  & 0.109  \\ 
  ti(Word Length, List) & 2.3725 & 3.3334 & 16.7243 & 0.0012 &       & 0.002  \\ 
   \hline
\end{tabular}
\caption{Summary of a Poisson GAM fitted to the total number of fixations (AIC: 70005.9).} 
\label{tab:fixcount}
\end{table}


%

\subsection{Naming latencies}

Two generalized additive models were fitted to the log-transformed naming latencies. Both models including an AR(1) process for the errors ($\rho=0.2$) and used the \texttt{scat} directive to adjust t-distributed residuals to normality.
Both models also included as control variables \textit{Manner} of articulation, \textit{Trial, List}, and \textit{Word Length}. 

\begin{figure}[htbp]
    \centering
    \includegraphics[width=0.90\textwidth]{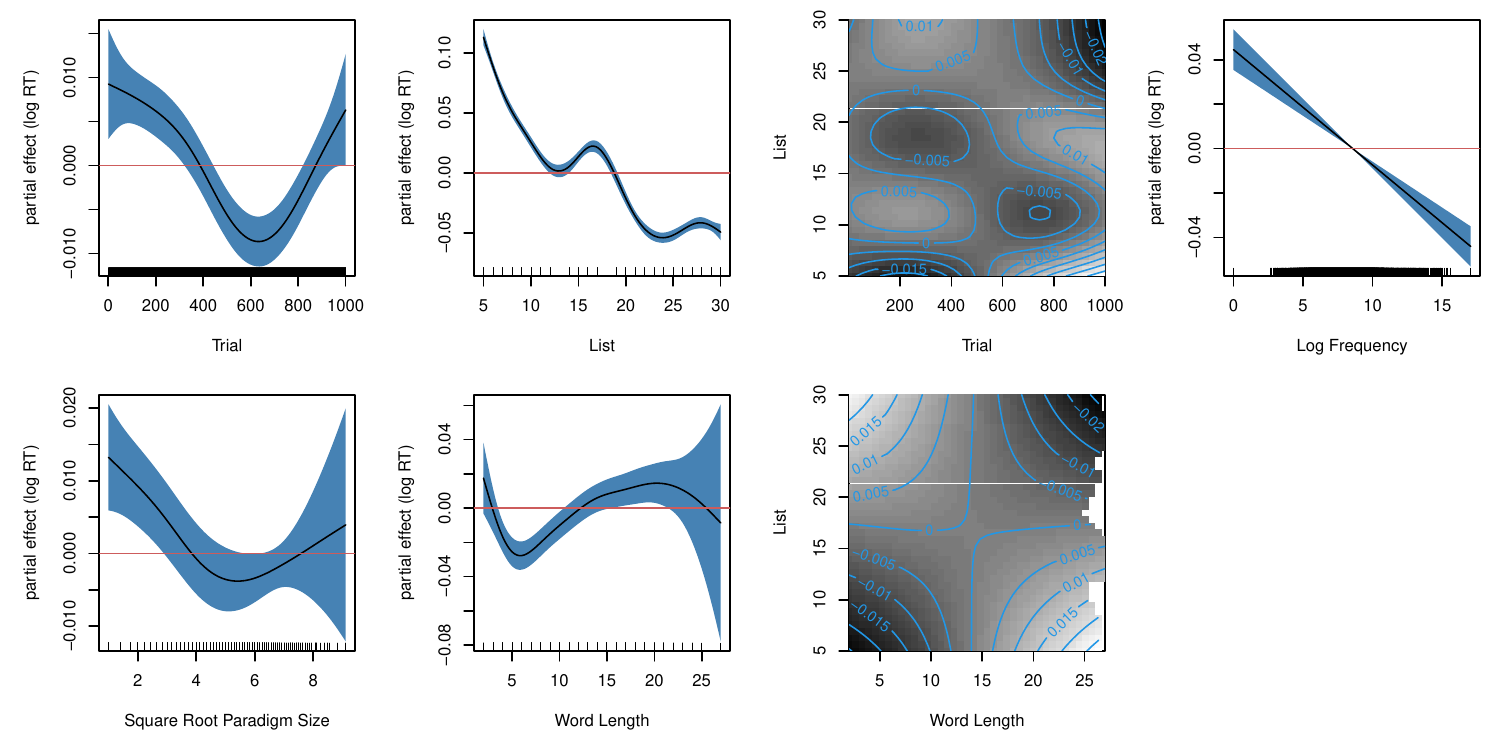}
    \caption{Partial effects of the smooths in a GAM fitted to log-transformed naming latencies, with control predictors and static lexicon predictors.}
    \label{fig:classical_RTs}
\end{figure}

The first model included further classical predictors: part of speech (\textit{POS}, a factor with 13 levels), the interpretational entropy of the word (\textit{Ent}), log-transformed) frequency of occurrence (\textit{Log Frequency}) and the (square-root transformed) number of inflected variants attested for a word's lemma (\textit{Paradigm Size}). Table~\ref{tab:RTclassical} provides the summary for this model, and Figure~\ref{fig:classical_RTs} presents plots for the nonlinear effects. The upper left panel of this figure shows that in the course of a session, naming latencies first decreased and then increased again. In the course of the sequence of sub-experiments, naming latencies generally decreased, but in an undulating way (second panel). The magnitude of these undulations varied somewhat by trial (third  panel). The upper right panel shows that naming latencies decreased linearly with frequency. The bottom center panel indicates that the lower values of paradigm size predict longer naming latencies. Shorter RTs for words with larger inflectional paradigms have been reported in earlier studies in Estonian \citep{Loo:etal:2018a,Loo:etal:2018b}. The second lower panel clarifies that both very short and longer words elicited longer reaction times, with an optimum around lengths 5--6. The final panel on the bottom row presents the interaction of word length by list.  For early lists, longer words elicited somewhat longer naming latencies, but by the end of the experiment, this effect reversed such that longer words were named slightly more quickly.

Using the increase in AIC when a predictor and all its interaction terms are withheld from the model specification as a measure of variable importance, \texttt{List} and \texttt{Manner} emerge as the strongest predictors, followed by \texttt{Word Length} and \texttt{Trial}. \texttt{POS} and \texttt{Entropy} have the lowest variable importance.

Table~\ref{tab:RTclassical} also lists the concurvity scores for the smooth terms. Concurvity scores are very small for \texttt{Trial} and the interaction of \texttt{Trial} by \texttt{List}.  For predictors of theoretical interest, concurvity scores are high, ranging from 0.49 for word frequency to 0.72 for inflectional paradigm size. This makes it difficult to assess the individual contributions of these predictors. For word frequency, for instance, almost 50\% of its effect is predictable from the other predictors in the model.  

\begin{figure}[htbp]
    \centering
    \includegraphics[width=0.65\textwidth]{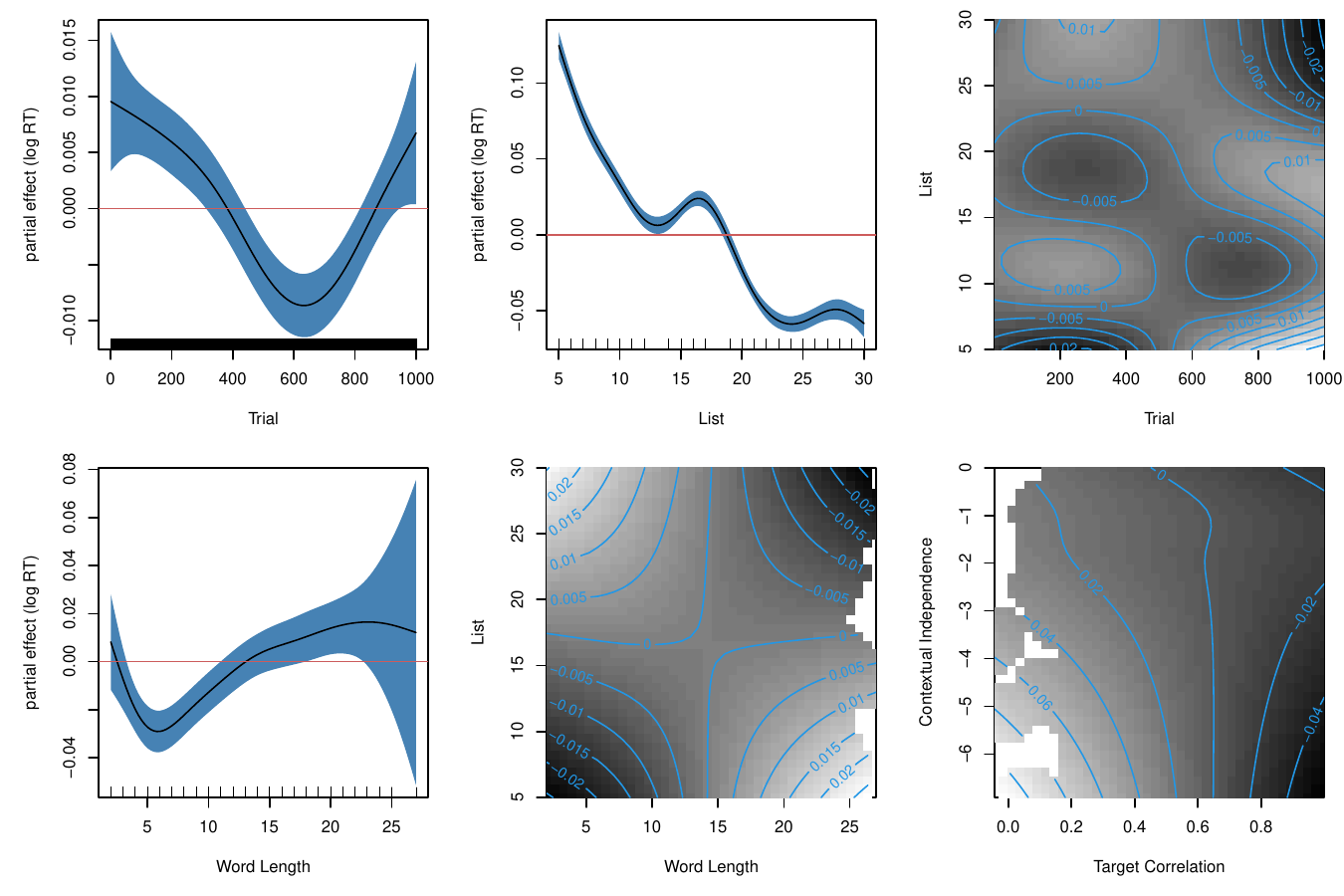}
    \caption{Partial effects of the smooths in a GAM fitted to log-transformed naming latencies, with FIDDL-based \textit{Target Correlation} and Contextual Independence (\textit{Cind}).}
    \label{fig:FIDDL_RTs}
\end{figure}

The second model replaced the paradigm size, frequency, part of speech, and entropy measures with \textit{Target Correlation} and Contextual Independence (\textit{Cind}).  Figure~\ref{fig:FIDDL_RTs} presents the partial effects of the smooth terms, and Table~\ref{tab:RT_FIDDL} presents the model summary.  A comparison of Figures~\ref{fig:classical_RTs} and \ref{fig:FIDDL_RTs} clarifies that the effects of \textit{Trial} and \textit{List} are estimated to be very similar by the two models. The same holds for the effect of \textit{Word Length}.  

The lower right panel of Figure~\ref{fig:FIDDL_RTs} presents the interaction of \textit{Target Correlation} and \textit{Cind}.  Naming latencies generally decreased with increasing \textit{Target Correlation}, but this effect becomes weaker for larger values of contextual independence.  For target correlations in the interval [0.2, 0.8], naming latencies decrease as \texttt{Cind} increases, for the remaining highest values of target correlation, the effect of \texttt{Cind} reverses with a small effect size.  Of the two  DLM models, the Target Correlation of the FIDDL model outperformed the corresponding measure from FIL model (AICs: -35041 and -34995 respectively), therefore Figure~\ref{fig:FIDDL_RTs} and Table~\ref{tab:RT_FIDDL} report the GAM with FIDDL-based Target Correlation. The variable importance of the interaction of Target Correlation by Contextual Independence is solid, and its concurvity score is low, less than half the concurvity of frequency in the classical model, and less than a third of that of Paradigm Size.  Therefore, the contribution of the learning-based predictors is less confounded with other predictors, and notably, word length.

The classical model has a lower AIC (-35077) than the FIDDL model (-35041), and  clearly is the better model. Given that the FIDDL model is trained on only the words in the experiment, this is perhaps unsurprising. Nevertheless, the amount of variance explained by the two models is very similar (30.3\% and 30.4\%). In the general discussion, we reflect on a variety of reasons that may help explain the advantage of the model with classical predictors.

\nopagebreak
\begin{table}[ht]
\centering
{\footnotesize
\begin{tabular}{lrrrrrr} \hline
A. parametric coefficients & Estimate & Std. Error & t-value & p-value & $\Delta$AIC & \\ 
  Intercept & -0.7123 & 0.0074 & -96.6021 & $<$ 0.0001 & & \\ 
  Manner: fricative & -0.1374 & 0.0023 & -58.4839 & $<$ 0.0001 & 3554 &\\ 
  Manner: nasal & -0.0212 & 0.0026 & -8.0833 & $<$ 0.0001 & \\ 
  Manner: plosive &0.0050 & 0.0019 & 2.6781 & 0.0074 & \\ 
  Manner: trill & -0.0135 & 0.0033 & -4.1269 & $<$ 0.0001 & \\ 
  Manner: vowel & -0.0329 & 0.0021 & -15.6505 & $<$ 0.0001 & \\ 
  Entropy: zero & -0.0100 & 0.0030 & -3.2740 & 0.0011 & 7 & \\ \hline
B. smooth terms & edf & Ref.df & F-value & p-value & $\Delta$AIC & concurvity\\ 
  s(Trial) & 4.0515 & 5.0013 & 11.4353 & $<$ 0.0001 & 218    & 0.01460 \\ 
  s(List) & 8.4289 & 8.9057 & 156.9420 & $<$ 0.0001 & 4661    & 0.9579 \\ 
  ti(Trial,List) & 11.1064 & 13.4305 & 7.6255 & $<$ 0.0001&      & 0.00798 \\ 
  s(Log Frequency) & 1.0009 & 1.0017 & 94.3852 & $<$ 0.0001& 95     & 0.49119 \\ 
  s(POS) & 5.0504 & 13.0000 & 2.4803 & $<$ 0.0001 &  22   & 0.53569 \\ 
  s(Paradigm Size) & 3.0206 & 3.8034 & 10.9678 & $<$ 0.0001 & 42    & 0.72334 \\ 
  s(Word Length) & 6.3555 & 7.2403 & 42.4383 & $<$ 0.0001 & 274    & 0.636640 \\ 
  ti(WordLength, List) & 1.9963 & 2.7806 & 15.7227 & $<$ 0.0001 &     & 0.46312 \\ \hline
\end{tabular}
}
\caption{Summary of a Gaussian GAM fitted to the log-transformed \textbf{naming latencies} with control predictors (\textit{Manner, Trial, List}) and \textbf{classical predictors} (log-transformed \textit{Frequency}, part of speech (\textit{POS}), inflectional \textit{Paradigm Size}, and \textit{Word Length}. The reference level for \textit{Manner} is `approximant', and for \textit{Entropy}, non-zero entropy. Concurvity scores are given for the smooth terms. AIC: -35076.69. $\Delta$AIC: the increase in AIC when a predictor, and all its interaction terms, are removed from the model specification.} 
\label{tab:RTclassical}
\end{table}

\clearpage
\nopagebreak

\begin{table}[ht]
\centering
{\footnotesize
\begin{tabular}{lrrrrrr} \hline
A. parametric coefficients & Estimate & Std. Error & t-value & p-value & $\Delta$AIC & \\ 
  Intercept & -0.7079 & 0.0044 & -161.4832 & $<$ 0.0001 &       & \\ 
  Manner: fricative & -0.1376 & 0.0024 & -58.5408 & $<$ 0.0001  &  3564    &\\  
  Manner: nasal & -0.0213 & 0.0026 & -8.1486 & $<$ 0.0001  &      &\\ 
  Manner: plosive & 0.0049 & 0.0019 & 2.6107 & 0.0090  &      &\\ 
  Manner: trill & -0.0125 & 0.0033 & -3.8192 & 0.0001  &      &\\ 
  Manner: vowel & -0.0337 & 0.0021 & -16.0192 & $<$ 0.0001  &      &\\ \hline
B. smooth terms & edf & Ref.df & F-value & p-value& $\Delta$AIC & concurvity \\ 
  s(Trial) & 4.0344 & 4.9810 & 11.3718 & $<$ 0.0001 & 213 &   0.0119     \\ 
  s(List) & 8.4338 & 8.9071 & 132.0102 & $<$ 0.0001 & 4639 & 0.9609    \\ 
  ti(Trial, List) & 10.8944 & 13.2264 & 7.5489 & $<$ 0.0001 &     \\ 
  s(Word Length) & 6.0759 & 6.9948 & 35.3161 & $<$ 0.0001 & 204 &  0.6842  \\ 
  ti(Word Length, List) & 2.3420 & 3.3099 & 14.5077 & $<$ 0.0001 & &      \\ 
  te(Target Correlation, Cind) & 6.8167 & 7.9454 & 24.7164 & $<$ 0.0001 & 176 & 0.2045    \\ 
   \hline
\end{tabular}
}
\caption{Summary of a GAM fitted to log naming latency with DLM predictors: \textit{Target Correlation} calculated from a FIDDL model, and Contextual Independence (\textit{Cind}). AIC: -35041. Concurvity scores are given for the smooth terms. $\Delta$AIC: the increase in AIC when a predictor, and all its interaction terms, are removed from the model specification.} 
\label{tab:RT_FIDDL}
\end{table}

\clearpage

\subsection{Word duration}

\begin{figure}[htbp]
    \centering
    \includegraphics[width=0.8\textwidth]{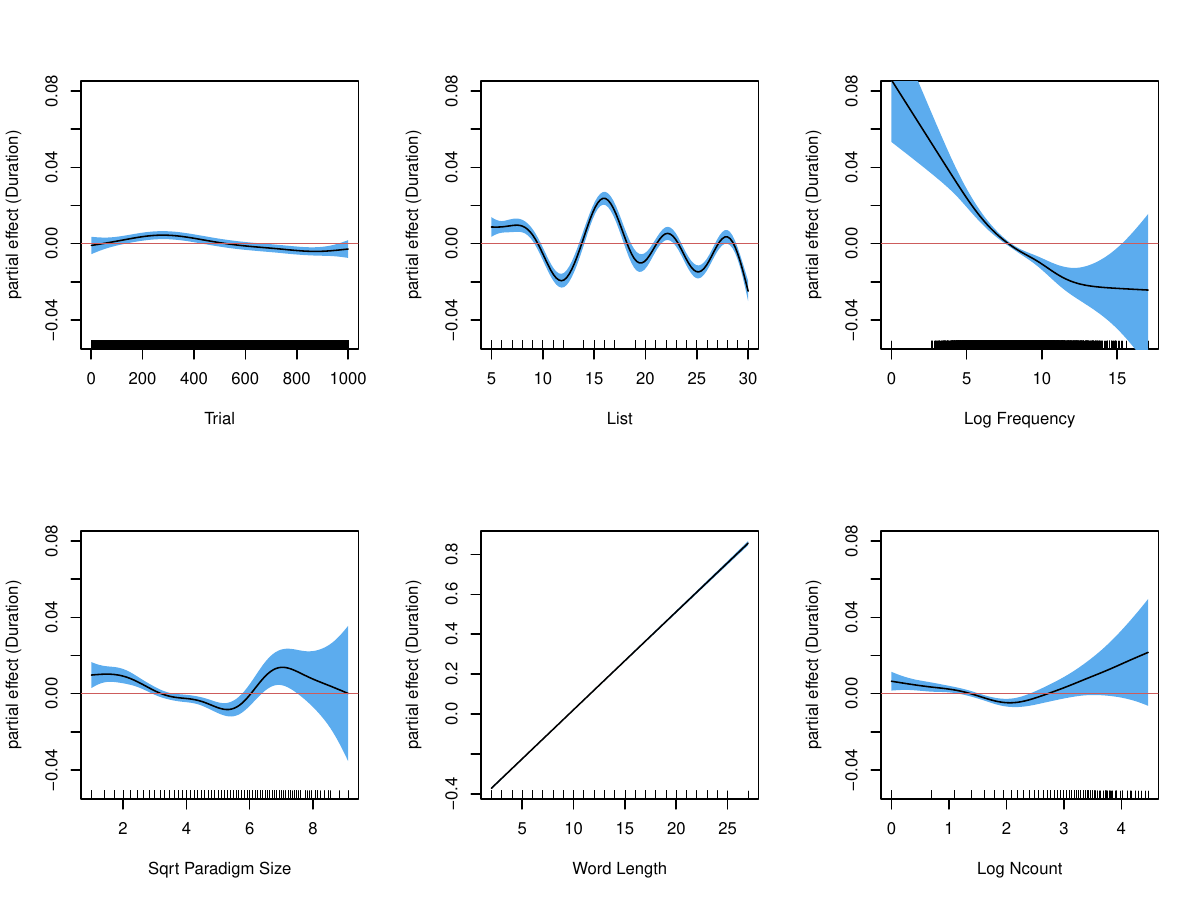}
    \caption{Partial effects of the smooths in the GAM model fitted to \textbf{spoken word duration} with \textbf{classical} predictors (AIC: -38281).}
    \label{fig:classical_duration}
\end{figure}

We also analyzed the spoken word durations, again comparing a GAM with classical predictors and a GAM with DLM-based predictors. Both GAMs set the family directive to \texttt{scat}, thus using a scaled t-distribution for the residuals. Figure~\ref{fig:classical_duration} and Table~\ref{tab:classical_duration} report on the model with the classical predictors.  A small effect of trial is visible, although it is not fully supported given that we set $\alpha=0.0001$. Unsurprisingly, its variable importance is small ($\Delta$ AIC = 18). An undulating effect of list is well-supported.  Word length was a strong predictor with a linear effect. Unsurprisingly, word length in seconds increased with word length measured in letters.  

Turning to the predictors of theoretical interest, as expected, spoken word durations decreased with frequency, but less so for higher frequencies.  The neighborhood count had a small, almost linear effect for values below 2.5.
For larger values, no effect was present.
Its variable importance is tiny ($\Delta$ AIC = 8). For the (square-root transformed) paradigm size, a U-shaped effect emerged. For the highest values, no effect was present. 

For this model, there is medium concurvity, with highest values for POS and word length.  Given the ubiquitous correlations between neighborhood counts, word length, word frequency, and paradigm size, these concurvity scores are unsurprising. Importantly, they clarify that in this model, it is not possible to clearly disentangle from each other the effects of the five theoretically motivated predictors.


The model with DLM predictors is summarized in Table~\ref{tab:FIL_duration_cut}.  Effects of control variables (\texttt{Trial, List}, and Word \texttt{Length}) are extremely similar to those in the model with classical predictors, and hence are not shown. 

\begin{figure}[htbp]
    \centering
    \includegraphics[width=0.4\textwidth]{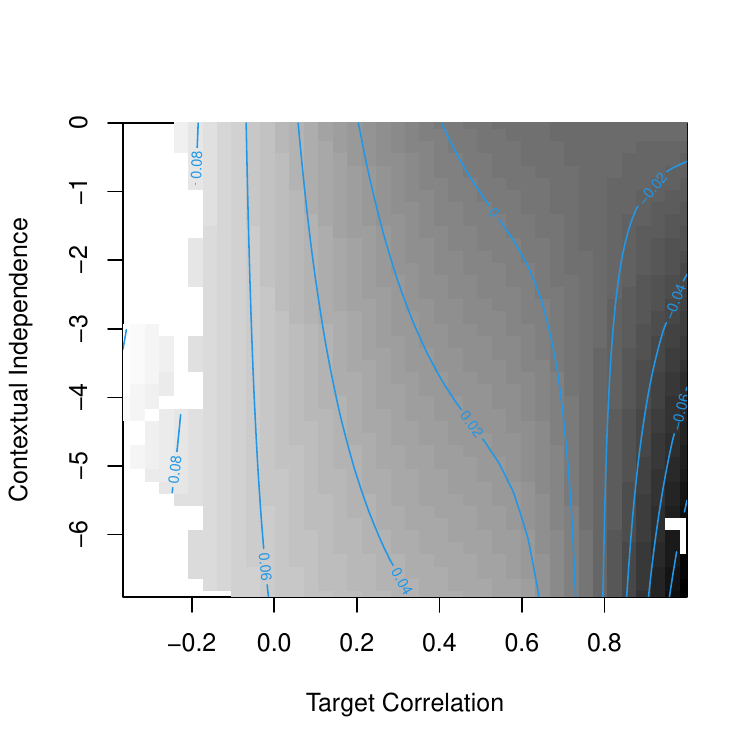}
    \caption{Partial effect of the tensor smooth of FIL-based \textbf{Target Correlation} and \texttt{Contextual Independence} in the GAM for spoken word duration.}
    \label{fig:FIL_duration_cut}
\end{figure}

As observed for the total fixation durations and naming latencies, an interaction of Target Correlation by Contextual Independence is present. The tensor product smooth for these predictors  is visualized in Figure~\ref{fig:FIL_duration_cut}. Overall, greater target correlations predict shorter word durations. For intermediate values of target correlation, a greater contextual independence predicts shorter durations, but this effect reverses for the largest values of target correlation. The effect of \texttt{Cind} is modest, but well supported: removal of this predictor from the model specification results in an increase of 70 AIC units. 

The model based on FIL (AIC -38200) outperformed the model based on FIDDL (-38049). Figure and table therefore report the FIL-based predictors.

It is noteworthy for the DLM model that all concurvity scores are small, and that the effects of \texttt{Contextual Independence} and \texttt{Target Correlation} do not show signs of being confounded with word length. Thus, although the model with classical predictors provides the better fit (AIC -38281) compared to the DLM model (AIC -38200), it offers a clearer window on the word durations. We note that in terms of the amount of variance explained, the two models perform very similar(73.6\% for the DLM model, 73.8\% for the classical model).

\begin{table}[ht]
\centering
{\footnotesize 
\begin{tabular}{lrrrrrr} \hline
A. parametric coefficients & Estimate & Std. Error & t-value & p-value & $\Delta$AIC &  \\ 
  Intercept & 0.6806 & 0.0075 & 90.9099 & $<$ 0.0001 &  & \\ 
  Entropy: zero & 0.0085 & 0.0030 & 2.8897 & 0.0039 & 8 & \\ \hline
B. smooth terms & edf & Ref.df & F-value & p-value & $\Delta$AIC &  concurvity\\ 
  s(trial) & 3.8203 & 4.7256 & 5.0712 & 0.0002 &  18          &   0.0037       \\ 
  s(listnum) & 8.8750 & 8.9952 & 43.6366 & $<$ 0.0001 & 384           &  0.0033        \\ 
  s(LogFormFreqENC23) & 3.7160 & 4.7130 & 31.9981 & $<$ 0.0001 &  135          & 0.2010 \\
  s(POS) & 6.2544 & 12.0000 & 10.7303 & $<$ 0.0001 &    93        &   0.5461  \\ 
  s(ParadSqrt) & 6.2990 & 7.2217 & 8.9295 & $<$ 0.0001 &  52          &  0.3691        \\ 
  s(wordLength) & 1.1185 & 1.2276 & 29856.2050 & $<$ 0.0001 & 19841           &  0.4254        \\ 
  s(LogNcount) & 3.9419 & 4.7710 & 5.8266 & $<$ 0.0001 &   8         &  0.3221        \\ \hline
\end{tabular}
}

\caption{Model summary of a Scaled t(5.274,0.074) GAM fitted to spoken word \textbf{duration}, using \textbf{classical} predictors (AIC:-38281).} 
\label{tab:classical_duration}
\end{table}

\clearpage

\begin{table}[ht]
\centering
{\footnotesize
\begin{tabular}{lrrrrrr} \hline
A. parametric coefficients & Estimate & Std. Error & t-value & p-value & $\Delta$AIC & \\ 
  Intercept & 0.6992 & 0.0006 & 1149.6326 & $<$ 0.0001 \\ \hline
B. smooth terms & edf & Ref.df & F-value & p-value& $\Delta$AIC & concurvity \\ 
  s(Trial) & 3.4710 & 4.3053 & 5.2713 & 0.0002 & 18 & 0.0025 \\ 
  s(List) & 8.8735 & 8.9951 & 44.0365 & $<$ 0.0001 & 391 & 0.0017 \\ 
  s(Word Length) & 2.5393 & 3.2269 & 19499.3995 & $<$ 0.0001 & 26174 & 0.0863 \\ 
  te(Target Correlation, Cind) & 7.1753 & 8.0686 & 48.0983 & $<$ 0.0001 & 366 & 0.0073 \\ \hline
\end{tabular}
}

\caption{Summary of the scaled t(5.129,0.073) GAM fit to spoken word \textbf{duration} with FIL-based \textbf{Target Correlation} and contextual independence (\textit{Cind}) as predictors (AIC: -38200).} 
\label{tab:FIL_duration_cut}
\end{table}


\clearpage

\section{General Discussion}

Table~\ref{tab:overview} provides an overview of the main predictors and whether they contribute to the models for the five response variables: first fixation duration, number of fixations, total fixation duration, naming latency, and spoken word duration.

Trial and list contributed without exception to each of the five models, indicating that the effects of experimental time, at both short time scales (trial within session) and long time scales (session) are pervasive. These longitudinal changes are likely due to fluctuations in attention, learning, and fatigue  \citep[see also][]{Baayen:Vasishth:Kliegl:Bates:2017,Pham:Baayen:2015,heitmeier2023trial}.

\begin{table}[hbp]
\centering
\begin{tabular}{lccccc} \hline
                               & FirstFixDur &  NFix  &  TotFixDur  &   RT    & Duration \\ \hline 
   Trial                       & $+$         &   $+$  &   $+$        &  $+$    &    $+$ \\
   List                        & $+$         &   $+$  &   $+$        &  $+$    &    $+$ \\ \hline
   Word Length                 & $+$         &   $+$  &   $+$        &  $+$    &    $+$ \\
   Word Length $\times$ List   & $-$         &   $(+)$&   $+$        &  $+$    &    $-$ \\ \hline
   Word Frequency              & $-$         &   $-$  &   $+$        &  $+$    &    $+$ \\
   Ncount                      & $-$         &   $-$  &   $-$        &  $-$    &    $+$ \\
   Infl. Family Size           & $-$         &   $-$  &   $-$        &  $+$    &    $+$ \\ \hline
   targetCorrelation FIL       & $-$         &   $-$  &   $+$        &  $+$    &    $+$ \\
   targetCorrelation FIDDL     & $-$         &   $-$  &   $+$        &  $+$    &    $+$ \\
   Contextual independence     & $-$         &   $-$  &   $+$        &  $+$    &    $+$ \\ \hline
\end{tabular}
\caption{Overview of supported effects of predictors by response variable.}\label{tab:overview} 
\end{table}

Just as trial and list, word length was always predictive. Interestingly, the effect of word length changed somewhat as the experiment progressed, as indicated by the interaction of word length by list, which was present for the total fixation durations and the naming latencies, and possibly also for the number of fixations. This interaction of frequency by length is likely due to the over-representation of longer words in the experiment. In normal language use, shorter words are used most often. In the experiment, however, high-frequency words are presented only once, and large numbers of longer lower frequency words are encountered at a rate that is atypical for normal language use.  The participant is adjusting to this atypical distribution of frequencies.  Later in the experiment, longer words required fewer fixations,  less long fixations, and  response latencies became somewhat shorter. In other words, in the course of the experiment, the participant became more skilled at reading out loud longer Estonian words. This may be due to a changing task strategy, to lexical learning in the course of the experiment, or both.  This finding provides a clear counterexample to the claim of \citet{thul2021using} that in experiments, longitudinal effects of experimental time do not interact with predictors of theoretical interest. \citet{baayen2022note} provided a counterexample from the British Lexicon Project \citep{Keuleers:etal:2012}. Here, we have shown that word length, a predictor of clear theoretical interest with a pervasive effect across all response variables, has an effect that changes in the course of the experiment. Furthermore, the interaction of word length by session is also of theoretical interest, as it shows that in a mega-experiment, participants can learn to adjust to the atypical distribution of words presented to them.

The classical standard predictors for lexical processing --- word frequency, neighborhood density, and inflectional paradigm size --- were predictive only for the temporal measures: total fixation duration, naming latencies, and spoken word durations.  Their irrelevance for predicting first fixation durations and the total number of fixations suggests that, at least for the present experiment, visual scanning and visual information uptake was not influenced by the participant's lexical knowledge.  This conclusion is supported by the fact that also the measures motivated by the DLM --- target correlation and contextual independence --- are not predictive for the first fixation duration or the number of fixations.  For the three temporal measures, they are consistently predictive. 

A problem present for many of the GAM models using classical predictors is substantial concurvity. For instance, in the case of the naming latencies, the concurvities range from 0.49 (for frequency) to 0.63 (for length), making it impossible to tell with precision what the contribution of frequency is, independently from word length, and vice versa. 

Problems of concurvity are much reduced when word frequency, inflectional family size, and neighborhood density are replaced by the interaction of two DLM predictors, target correlation and contextual independence. Their joint contribution is much less confounded with word length than is the case for measures such as word frequency and neighborhood density.

\begin{table}[hbp]
\centering
\begin{tabular}{lrrr} \hline
              & TotFixDur  &       RT   &   Duration \\ \hline
   classical  &    271584  &   -35077   &     -38281 \\
   FIL        &    271585  &   -34995   &     -38200 \\
   FIDDL      &    271589  &   -35041   &     -38049 \\ \hline
\end{tabular}
\caption{Overview of model fit, using AIC, for the classical and DLM models, for the three temporal response variables (total fixation duration, response latency, and spoken word duration.}\label{tab:summary_AIC}
\end{table}

Table~\ref{tab:summary_AIC} provides an overview of model quality, using AIC, cross-classified by model type (classical, FIL, FIDDL) and temporal measure (total fixation duration, naming latency, and spoken word duration).  First note that for the total fixation duration, the FIL-based GAM is of the same quality as the GAM with classical predictors. The FIDDL-based GAM is performing slightly less well, but the difference in AIC is small. Conversely, for the naming latencies and the spoken word durations, the GAMs with classical predictors have a clear advantage over the FIL and FIDDL-based GAMs. Nevertheless, the models with the DLM predictors provide high-quality fits. When comparing the amount of variance explained, differences are in the order of magnitude of 1\%.  We note here that, because the DLM works with mappings from form to meaning (operationalized with embeddings), the predictivity of specifically the target correlation measure is a strong indication of the word naming task tapping into meaning.

The fact that the DLM-based GAMs perform as well as they do is surprising given that there are many constraints on the computational models that are cognitively suboptimal. First, the embeddings are based on FastText, which is trained on a text base that is much larger than what the participant in our experiment can ever have experienced. In other words, we have been pitting the embeddings of an AI against a single speaker, albeit a highly educated member of the Estonian language community.  

Second, the DLM models are trained on only the words in the dataset, which inevitably skews learning as large numbers of both common and uncommon words are not taken into account. For instance, whereas speakers have a relatively dense experience with the inflectional variants of high-frequency words, the model has to learn from much sparser data \citep[for further discussion, see][]{Nikolaev:Chuang:Baayen:2024}. This problem can be mitigated to some extent by following the training regime used by \citet{heitmeier2023trial}, and first train a model on a representative word frequency list (or corpus), and then take the model trial-by-trial through the experiment.   Yet, just as with the embeddings, the frequency norms taken from a corpus represent community knowledge, and offer only a rough approximation of any individual speaker's vocabulary knowledge. 

Third, the DLM is a model of subliminal lexical statistical learning, and thus can capture only part of the cognitive processes that unfold during the naming task. For instance, some words are pronounced differently depending on how their case and number endings are interpreted. This is something the participant reported to be very aware of, yet this is not taken into account by the DLM measures. Metalinguistic knowledge that participants use in the naming task is outside the scope of the DLM.

Of theoretical interest is the finding that the model using a simple linear mapping (FIL) outperforms a model using a deep mapping (FIDDL) for two of the three temporal response variables (total fixation duration and spoken word duration).  It should be kept in mind that we have not explored the space of hyper-parameters for deep learning models, and have simply used the default settings suggested by \citet{heitmeierEtAl2024}. Nevertheless, it is surprising that the FIDDL model, with 1.5 billion connection weights, does not outperform FIL (with 1.5 million connection weights) when it comes to predicting aspects of Estonian lexical processing in the naming task. Recall that prediction accuracy on the models' training data, evaluated by asking whether the closest semantic neighbor of a word's predicted embedding is the embedding given by FastText for that word, is 74.4\% for FIDDL and 17.4\% for FIL.  This suggests that the deep model is over-trained on the limited set of words that it has to learn, and that accuracy on training data is not necessarily a precondition for obtaining accurate predictions for human lexical processing. 

The two DLM measures that we have used in this study decompose the standard word frequency effect into two conceptually very different components. The first component assesses how frequency of use affects the learning of the mapping between word form and meaning. The target correlation measure quantifies this component by considering how precisely a word's meaning has been learned.  The second component assesses how the use of words in utterances affects word knowledge.  As measures of in-context surprisal \citep[see, e.g.,][]{Levy:2008} are not useful for isolated word naming, we used a measure that evaluates to what extent a word can be understood on its own, in the absence of the words in its context that normally guide semantic interpretation. Following \citet{Gahl:Baayen:2024}, we therefore calculated words' contextual independence.   For total fixation duration, naming latency, and spoken word duration, Target Correlation and Contextual Independence interacted, in roughly the same way, with the effect of one predictor being greatest for small values of the other predictor.  The fact that these two measures are effective predictors of those response variables for which frequency, inflectional paradigm size, and neighborhood density (in varying combinations) are also predictive indicate that theories that take learning into account are now computationally tractable.\footnote{ 
It has been pointed out repeatedly that the human mind cannot perceive the world as such, and that our perceptions are filtered through the specifics of human perceptual system and human culture, see, e.g., \citet{kant1999critique,merleau2013phenomenology,Husserl:1913} and more recently,   \citet{hoffman2019case}.  Learning-driven models make it possible to take these insights into account, and to move away from the naive realism of theories that posit abstract fixed representations for words and counters in the head that keep track of the supposedly veridical observable occurrences of these words in the environment.
}
 
Having acknowledged the limitations of computational modeling, it is nevertheless remarkable how competitive the GAM models with DLM predictors are, both in terms of AIC and proportion of variance explained, and in terms of interpretational clarity (reduced concurvity). We are not aware of any computational implementation of an interactive activation model that can handle languages with rich and complex morphologies such as Estonian, although the PONG model of \citet{Snell:2024} may prove us wrong. The problem as we see it is that the vast numbers of different word forms in languages such as Estonian would, in interactive activation models, require exponential amounts of between-word inhibitory connections, which renders them cognitively, biologically, and computationally unattractive. We hasten to add that we do not ascribe biological reality to the simple artificial neural networks that we have implemented. The strength of the DLM is that it offers tools for assessing in a mathematically precise way, the consequences of usage for how accurately speakers can learn the vocabulary of their language.  In the age of deep learning and large language models, the DLM models may look extremely simplistic, but they are surprisingly powerful.

\newpage

\section*{Acknowledgements}
This research was supported by Estonian Research Council grant number PSG743, awarded to the first author. The authors would like to thank Fabian Tomaschek for his help with the audio data preprocessing.

\newpage

\bibliography{data,library}
\end{document}